\documentclass[journal]{IEEEtran}

\usepackage{amsthm}
\usepackage{amsmath}
\usepackage{mathrsfs}
\usepackage{stmaryrd}
\usepackage{graphics,epsfig}
\usepackage{epstopdf}
\usepackage[table]{xcolor}
\usepackage{array,color}
\usepackage{verbatim}
\usepackage[tight,footnotesize]{subfigure}
\usepackage{booktabs}
\usepackage{ctable}
\usepackage{multirow}
\usepackage{morefloats}
\usepackage[ruled,vlined,linesnumbered]{algorithm2e}
\usepackage{cite}

\usepackage{graphicx}
\usepackage{graphics,epsfig}
\usepackage{epstopdf}
\usepackage{amssymb}
\usepackage{amsmath}
\usepackage[footnotesize]{subfigure}
\usepackage{verbatim}
\usepackage{booktabs}
\usepackage{multirow}
\usepackage{morefloats}
\usepackage{cite}
\usepackage{stfloats}
\usepackage{multirow}

\ifCLASSINFOpdf
\else
\fi

\begin{document}
\title{Facial Feature Embedded CycleGAN \\ for VIS-NIR Translation}
%



\author{\IEEEauthorblockN{Huijiao~Wang, Li~Wang, Xulei~Yang, Lei~Yu, and Haijian~Zhang}
\thanks{Huijiao~Wang, Lei~Yu, and Haijian~Zhang are with Signal Processing Laboratory, School of Electronic Information, Wuhan University, China. Li~Wang and Xulei~Yang are with Agency for Science, Technology and Research, Singapore.}
}

%

\maketitle

\begin{abstract}
VIS-NIR face recognition remains a challenging task due to the distinction between spectral components of two modalities and insufficient paired training data. Inspired by the CycleGAN, this paper presents a method aiming to translate VIS face images into fake NIR images whose distributions are intended to approximate those of true NIR images, which is achieved by proposing a new facial feature embedded CycleGAN. Firstly, to learn the particular feature of NIR domain while preserving common facial representation between VIS and NIR domains, we employ a general facial feature extractor (FFE) to replace the encoder in the original generator of CycleGAN. For implementing the facial feature extractor, herein the MobileFaceNet is pretrained on a VIS face database, and is able to extract effective features. Secondly, the domain-invariant feature learning is enhanced by considering a new pixel consistency loss. Lastly, we establish a new WHU VIS-NIR database which varies in face rotation and expressions to enrich the training data. Experimental results on the Oulu-CASIA NIR-VIS database and the WHU VIS-NIR database show that the proposed FFE-based CycleGAN (FFE-CycleGAN) outperforms state-of-the-art VIS-NIR face recognition methods and achieves 96.5\% accuracy.
\end{abstract}


\IEEEpeerreviewmaketitle

\section{Introduction}
\IEEEPARstart{T}{}he development of deep neural networks and large-scale visible light (VIS) face images yields high accuracy in face recognition under well-controlled lighting condition \cite{Schroff2015FaceNet,Sheng2018MobileFaceNets}. However, face recognition in low lighting condition is still a very challenging problem, e.g., the near infrared (NIR) image based face recognition \cite{ouyang2016survey}. In recent years, VIS-NIR heterogeneous face recognition has attracted more and more attention from researchers and enterprises due to its practical value in many real-world applications, e.g., security surveillance and E-passport \cite{he2017CVPRlearning,gao2017semi}. 

The main issue in dealing with NIR images is due to the significant distinction between spectral components of VIS and NIR images, as NIR images are insensitive to VIS illumination variations \cite{Zhu2017Matching}. Furthermore, for supervised learning \cite{2018AAAiADHFR}, it is not easy to collect pairwise VIS-NIR face images partially due to individual privacy. Therefore, this motivates us to design an effective network architecture without requiring a large number of pairwise face images for training.  

\begin{figure}[t]
\begin{center}
\includegraphics[width=1\linewidth]{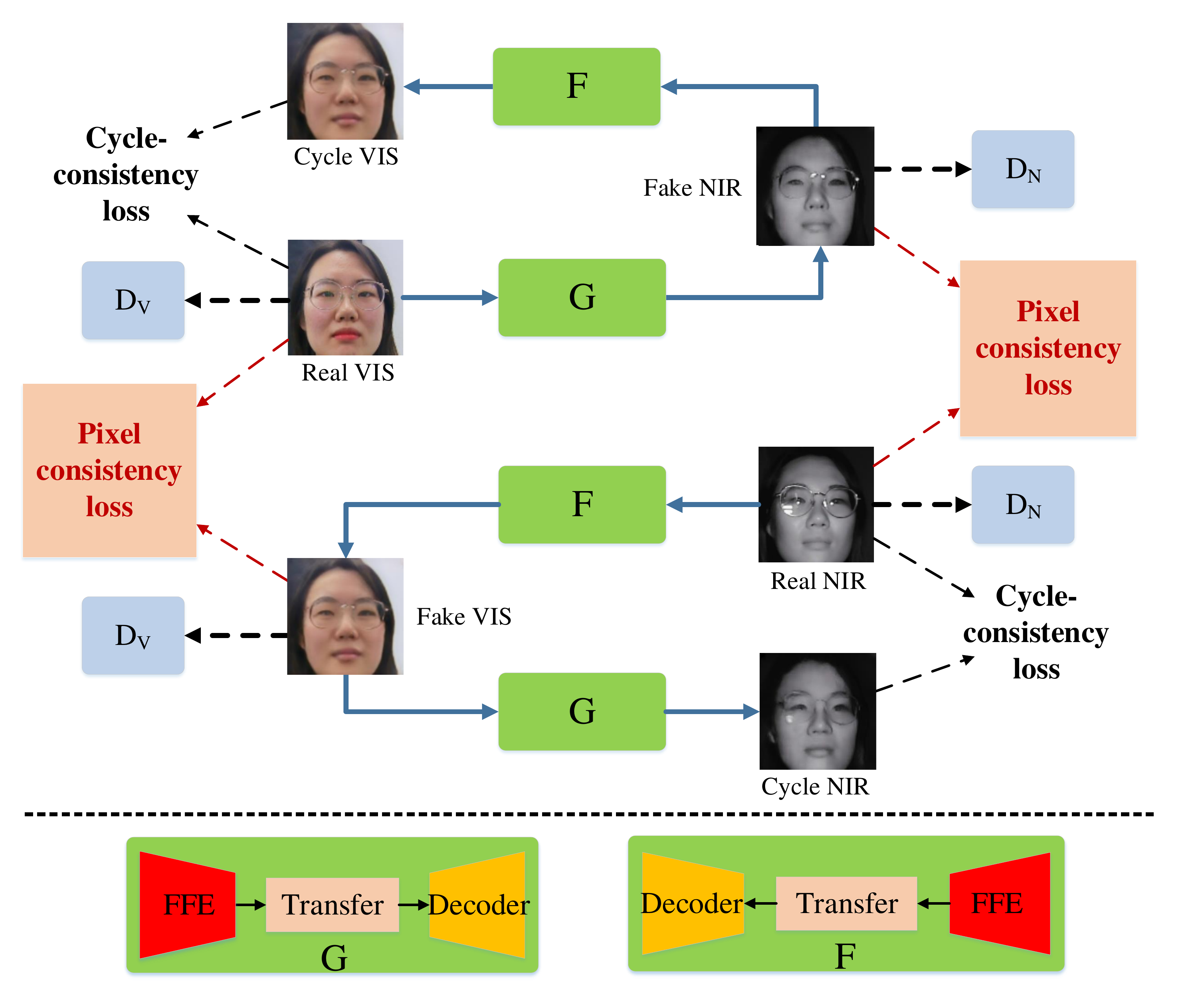}
\end{center}
\caption{Flowchart of the FFE-CycleGAN: a facial feature extractor (FFE) and a deconvolution module (Decoder) are embedded  in the two generotors (G and F) of the original CycleGAN \cite{Zhu2017Unpaired} to extract features from face images efficiently. Besides, we propose a pixel-consistency loss based on the cycle-consistency and adversarial losses in purpose of constraining the generated NIR images.}
\label{fig:long}
\label{fig:onecol}
\end{figure} 

This paper presents a new unpaired VIS-NIR translation method, named facial feature extractor based CycleGAN (FFE-CycleGAN), which can generate NIR face hallucination having similar spectral components as those of true NIR face images. The flowchart of the FFE-CycleGAN is drawn in Figure 1. Our experiments show that the original CycleGAN cannot obtain valid translation results between VIS and NIR face images (see Figure 5 and 6). To address this problem, we propose to use a general face feature extractor (e.g., the MobileFaceNet \cite{Sheng2018MobileFaceNets}) to replace the encoder in the generator of CycleGAN \cite{Zhu2017Unpaired}. To well preserve the facial feature, we propose a pixel consistency loss to measure similarities between hallucinations and true face images. Another contribution in this paper is to establish a so-called WHU VIS-NIR paired face database, which consists of some extreme facial rotations and makes the network more stable for transferring images. The proposed FFE-CycleGAN is evaluated on both WHU VIS-NIR database and well-known Oulu-CASIA NIR-VIS facial expression database \cite{2011PR_oulu-dataset}. Superior performance is achieved compared to the existing state-of-the-art methods \cite{2009cvprLearningMappings,2009CVPR_Coupled,2013TPAMI_HFR,2013TIP_KDSR,2017TNNLSCross,he2017CVPRlearning,2018AAAiADHFR,Zhu2017Unpaired}.

The remainder of the paper is organized as follows: Section II reviews the related work in the literature. In Section III, we firstly present the proposed network architecture, and then elaborate the proposed pixel consistency loss. Section IV introduces datasets and protocols, and Section V provides experimental results on two databases for evaluating the FFE-CycleGAN. Finally, Section VI concludes the paper.


\section{Related Work}
  
Features extracted respectively from VIS and NIR domains exist a significant gap, although they belong to the same subject. This makes the study of VIS-NIR heterogeneous face recognition necessary in certain specific scenarios \cite{2017CVPRNotAfraidofDark}, e.g., in poor lighting conditions. A family of methods are reported to fill the gap between VIS and NIR data in common latent space, pixel space, feature space, and etc.

\begin{itemize}

\item \textbf{Common latent space based methods}: One of the commonly used strategies is to find mapping functions between the NIR and VIS domains.  The authors in \cite{2015CVPR_NIR-VIS} propose a dictionary learning approach, which is based on cross-spectral joint $\ell_{0}$ minimization to learn a mapping function between VIS and NIR modalities, and then the VIS images are reconstructed in the NIR domain and vice versa. 

\item \textbf{Pixel space based methods}: Methods focusing on pixel space are also called synthesis based methods. This kind of method tries to synthesize or transform face images from NIR domain to VIS domain and vice versa. The authors in \cite{Wang2009Analysis-by-Synthesis} propose an analysis-by-synthesis framework, called face analogy, which could transform face images between NIR and VIS. In \cite{2009cvprLearningMappings}, a method is proposed for synthesizing VIS images from NIR domain by learning mapping between different spectra images to reduce inter-spectral difference.   
  
\item \textbf{Feature space based methods}: Feature space based methods learn modality-invariant features by a shared NIR-VIS layer or using NIR images to fine-tune the VIS network, e.g., the authors in \cite{2016ICB_Transferring, 2016ECCV_HFRwithCNN} use NIR images to fine-tune the VIS deep networks and explore different metric learning strategies to reduce discrepancy between different modalities. The work in \cite{2016cvpr_Seeing} trains VisNet and NirNet to couple their output features by using a siamese network. By doing so, they can learn the relationship between cross-modal face images. In \cite{he2017CVPRlearning}, one network is proposed to map both NIR and VIS to a compact Euclidean space. This high-layer could simultaneously learn modality-invariant features and modality-variant spectrum information by two orthogonal subspaces. In \cite{2017arxiv_Coupled,Li2016ACM_Mutual}, a generative model is proposed for the process of generating face images in different modalities and an Expectation Maximization (EM) algorithm is designed to iteratively learn the model parameters. Moreover, this generative model is able to infer the mutual components.

\item \textbf{Pixel and feature based methods}: Recently, there are some methods taking both pixel space and feature space into consideration, aiming at maximizing discriminations among different subjects \cite{2018AAAiADHFR}. It becomes necessary to consider this discrimination especially when the number of subjects increases. To optimize a VIS deep model for cross-spectral face recognition, the method in \cite{2017CVPRNotAfraidofDark} consists of two core components, i.e., cross-spectral hallucination and low-rank embedding. Benefitting from the capability of  fitting data distribution and style transferring \cite{Goodfellow2014NIPSGAN,Zhu2017Unpaired}, the authors in \cite{2018AAAiADHFR} use the work in \cite{Zhu2017Unpaired} to produce cross-spectral face hallucination. They also integrate the hallucination network and discriminative feature learning into an end-to-end adversarial network. 
  
\end{itemize}

Our method is related to the existing works in \cite{Goodfellow2014NIPSGAN, Zhu2017Unpaired,2018AAAiADHFR}, which perform well in image style transferring and image generation. Inspired by these works, this paper improves the feature generator in the original CycleGAN  \cite{Zhu2017Unpaired} and proposes an objective function with pixel consistency loss, aiming at generating high-fidelity NIR face images for VIS-NIR face recognition. 


\section{The Proposed FFE-CycleGAN}

Generative image modeling has made much progress recently, and its outstanding performance in fitting data distribution, e.g., style transferring \cite{Zhu2017Unpaired, 2017Image-to-Image, Goodfellow2014NIPSGAN}, has attracted much research attention. Especially, a semisupervised image translation model, i.e., CycleGAN \cite{Zhu2017Unpaired} , has been designed and it works well in style transferring. 
  
Thanks to the CycleGAN, we can easily train a model with unpaired face images from NIR and VIS spectra. It is a great progress in solving the challenging problem of lacking paired datasets in the style transfer task. Inspired by the CycleGAN model, we initially employ this adversarial model to learn the style features of NIR modality. However, experimental results show that the original CycleGAN performs not well in transferring face attributes. A reasonable explanation is that the feature learning module in the CycleGAN is not specifically designed for face images. In order to generate high-fidelity NIR images, the FFE-CycleGAN model in Figure 1 is proposed to learn face feature effectively  in both VIS and NIR modalities. Additionally, a new loss is added to improve qualities of generated images, with the purpose of making the distribution of generated fake images close to the one of real images. In the following two subsections, the proposed network architecture is firstly introduced, and then a new pixel consistency loss function is defined for synthesizing high-quality NIR face images.                                                                                                                                                                                                                                                                                                                                                                                                                                                                                                                                                                                                                                                                                                                                                                                                                                                                                                                                                                                                                                                                                                                                                                                                                                                                                                                                                                                                                                                                                                                                                                                                                                                                                                                                                                                                                                                                                                                                                                                                                                                                                                                                                                                                                                                                                                                                                                                                                                                                                                                                                                                                                                                                                                                                                                                                                                                                                                                                                                                                                                                                                                                                                                                                                                                                                                                                                                                                                                                                                                                                                                                                                                                                                                                                                                                                                                                                                                                                                                                                                                                                                                                                                                                                      

\subsection{Network Architecture}


The flowchart of the facial feature embedded CycleGAN is shown in Figure 1, and the network architecture is depicted in Figure 2 with detailed analysis of the general face feature extractor (FFE). Note that the generator of the original CycleGAN consists of encoder, translate module, and decoder. The encoder can be seen as a feature extracting module, which encodes an input image into its corresponding  feature. The translate module converts the extracted feature from VIS to NIR domain, and the decoder synthesizes a fake NIR image from the NIR feature.
 
As illustrated in Figure 2, the FFE-CycleGAN model has two mappings $ G : I_{V} \rightarrow I_{N} $ and $ F : I_{N} \rightarrow I_{V} $. Specifically, G mapping tries to convert VIS images  $ I_{V} $ to NIR images $ I_{N} $, and F mapping converts NIR images $ I_{N} $ back to VIS images  $ I_{V} $. The G and F mapping functions include adversarial discriminator $ D_{V} $ and $ D_{N} $, where $ D_{V} $ and $ D_{N} $ aim to distinguish between real images and generated images. Due to limited paired VIS-NIR images, we pretrain a facial feature extractor on a large number of VIS face images for extracting face feature. Then, the encoder module is replaced by the pretrained facial feature extractor and subsequently conducts fine-tuning on VIS-NIR datasets.

\begin{figure}[t]
\begin{center}
   \includegraphics[width=0.9999\linewidth]{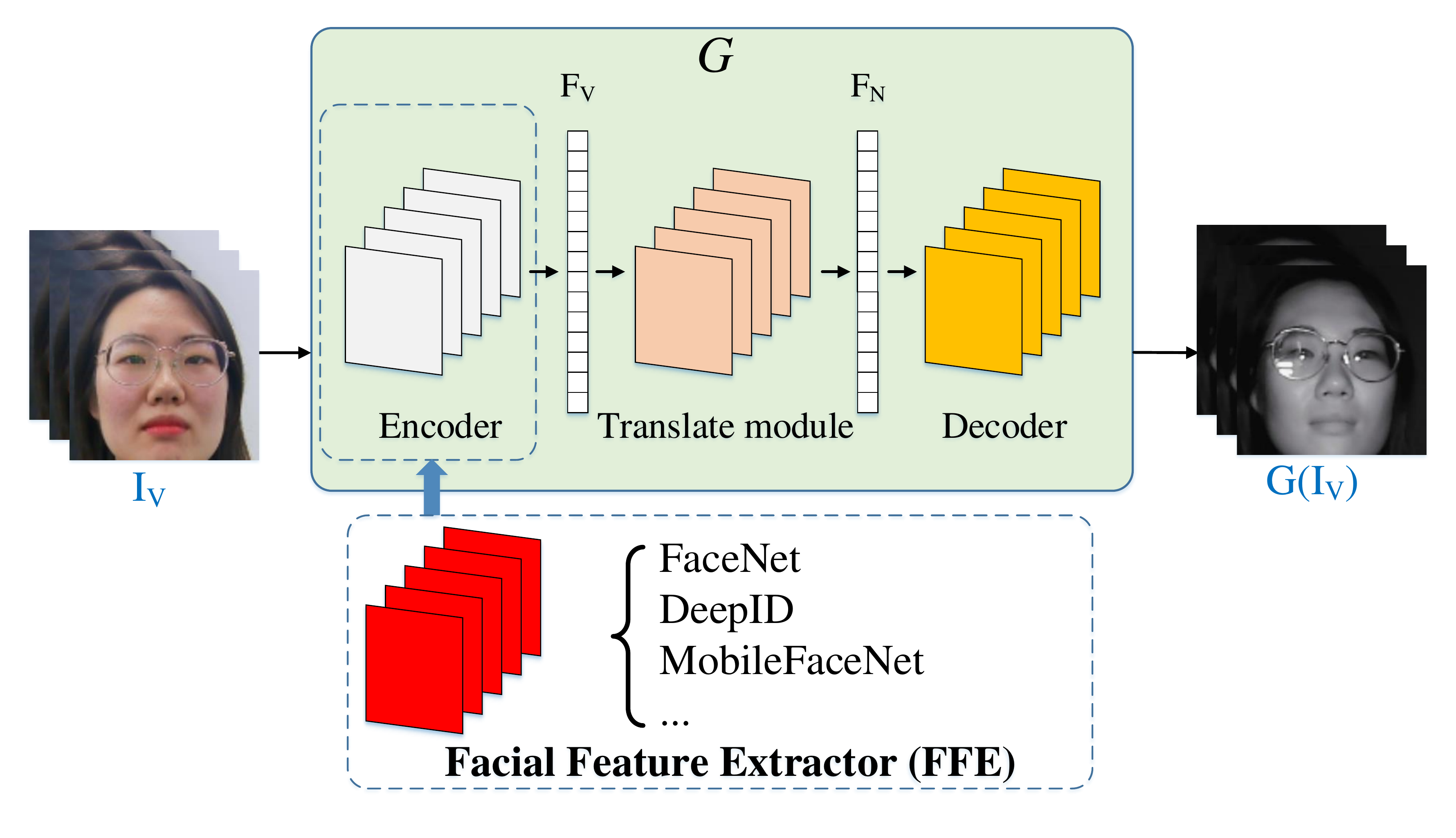}
\end{center}
   \caption{Structure of the generator G in FFE-CycleGAN: the pretrained FFE module is used to extract face feature ($F_{V}$) from true VIS images ($I_{V}$), which is better than the feature extracted from the original encoder of CycleGAN. Then,  $F_{V}$ is transferred by using 6 residual blocks to NIR image features $F_{N}$. Lastly, the NIR image  ($G(I_{V})$ ) is recovered by doing deconvolution (Decoder) on $F_{N}$ (The generator F in Figure 1 has the same structure as G).}

\label{fig:long}
\label{fig:onecol}
\end{figure} 

It is worth emphasizing that the capability of the FFE has significant influence on image transferring
results. In the following, we have reviewed some typical research works related to the facial feature extraction. 

\begin{itemize}

\item \textbf{FaceNet} \cite{Schroff2015FaceNet} directly trains a deep convolution network using triplets of roughly aligned matching/non-matching face patches. It maps face images directly to a compact Euclidean space to measure face similarities. It outperforms a lot of face feature extraction networks, and the recognition accuracy on LFW dataset is 99.63\%.

 \item \textbf{DeepID3} \cite{2015DeepID3} rebuilts two architectures from stacked convolution and inception layers proposed in VGG Net and GoogLeNet. They add joint face identification-verification supervisory signals to both intermediate and final feature extraction layers during training phase, and improve face recognition performance. These two architectures achieve 99.53\% face verification accuracy on LFW dataset. 

\item \textbf{MobileFaceNet} \cite{Sheng2018MobileFaceNets} is an extremely efficient CNN model with size of 4.0 MB, and uses less than 1 million parameters but has comparative performance as some state-of-the-art deep CNN models with size of hundreds of MB. It uses a global depthwise convolution (GDConv) \cite{Sandler2018MobileNetV2} as the global operator instead of average pooling, aiming at learning the importance of different spatial positions after training. Its face verification accuracy on LFW dataset can reach 99.55\%.  

\end{itemize}

Considering training complexity and verification accuracy, the MobileFaceNet model \cite{Sheng2018MobileFaceNets} is chosen to implement our face feature extractor. To generate  NIR face images that keep common latent facial features, this model is pretrained using MS-Celeb-1M dataset \cite{Guo2016ECCvMSCeleb1M}. As a result, it can effectively extract facial feature. 

Embedding the MobileFaceNet into the generator of CycleGAN is useful to learn face features. Instead of converting the feature extracted from a holistic input image, the translate module only converts the facial feature extracted by the FFE, which efficiently addresses the problem of transferring face style. So we need a face feature extractor which has the ability to extract face feature precisely. However, the number of NIR images is far less than that of VIS images. The huge difference of training sample amount between two modalities is another big challenge for the style transfer task.   Under such situation, it is difficult to train the FFE module and the whole generator well at the same time only on a small number of paired VIS-NIR images. Hence, we pre-train the FFE module taking full advantage of easy-to-get VIS images. The pre-trained FFE module can precisely  extract face features whatever the images are from VIS domain or NIR domain. Then, we train the FFE-CycleGAN network on paired VIS-NIR dataset to learn the mapping function between two domain. By doing so, we can synthesis more realistic NIR images.  

\subsection{Pixel Consistency Loss}
Following the pioneering work in \cite{Zhu2017Unpaired, Goodfellow2014NIPSGAN}, we replace the negetive log likelihood objective with a least square loss \cite{mao2016multi} as same as in \cite{Zhu2017Unpaired}. The adversarial losses for the mapping function $ G : I_{V} \rightarrow I_{N} $ and its discriminator $ D_{N} $ are formulated as follows:
\begin {equation}
\begin {aligned} 
\mathcal{L}_{\mathrm{LSGAN}}(G, D_{N}, & I_{V}, I_{N}) =\mathbb{E}_{i_{N} \sim P(i_{N})}\left[ (D_{N}(i_{N})-1)^{2}\right] \\ &+\mathbb{E}_{i_{V} \sim P(i_{V})}\left[ \left( D_{N}(G(i_{V}))^{2}\right]\right.,
\end {aligned} 
\end {equation}
 
where $ i_{V} $  and $ i_{N} $ are sample images from VIS domain $ I_{V} $  and NIR domain $I_{N} $ , respectively. $ G $ tries to minimize the objective while $ D_{N} $ aims to maximize it. The objective for $F$ mapping is similar to $G$ mapping being  expressed as $ \mathcal{L}_{\text{LSGAN}}(F, D_{V}, I_{N}, I_{V})$. The cycle consistency loss (cyc) is formulated as follows: 
\begin {equation}
\begin {aligned}
\mathcal{L}_{cyc}(G, F) = &\mathbb{E}_{i_{V} \sim P(i_{V})}\bigg[||F(G(i_{V}))-i_{V}||_1\bigg]  \\
 &+ \mathbb{E}_{i_{N} \sim  P(i_{N})}\bigg[||G(F(i_{N}))-i_{N}||_1\bigg].
 \end {aligned}
\end {equation}

     \begin{figure}[t]
\begin{center}
   \includegraphics[width=0.95\linewidth]{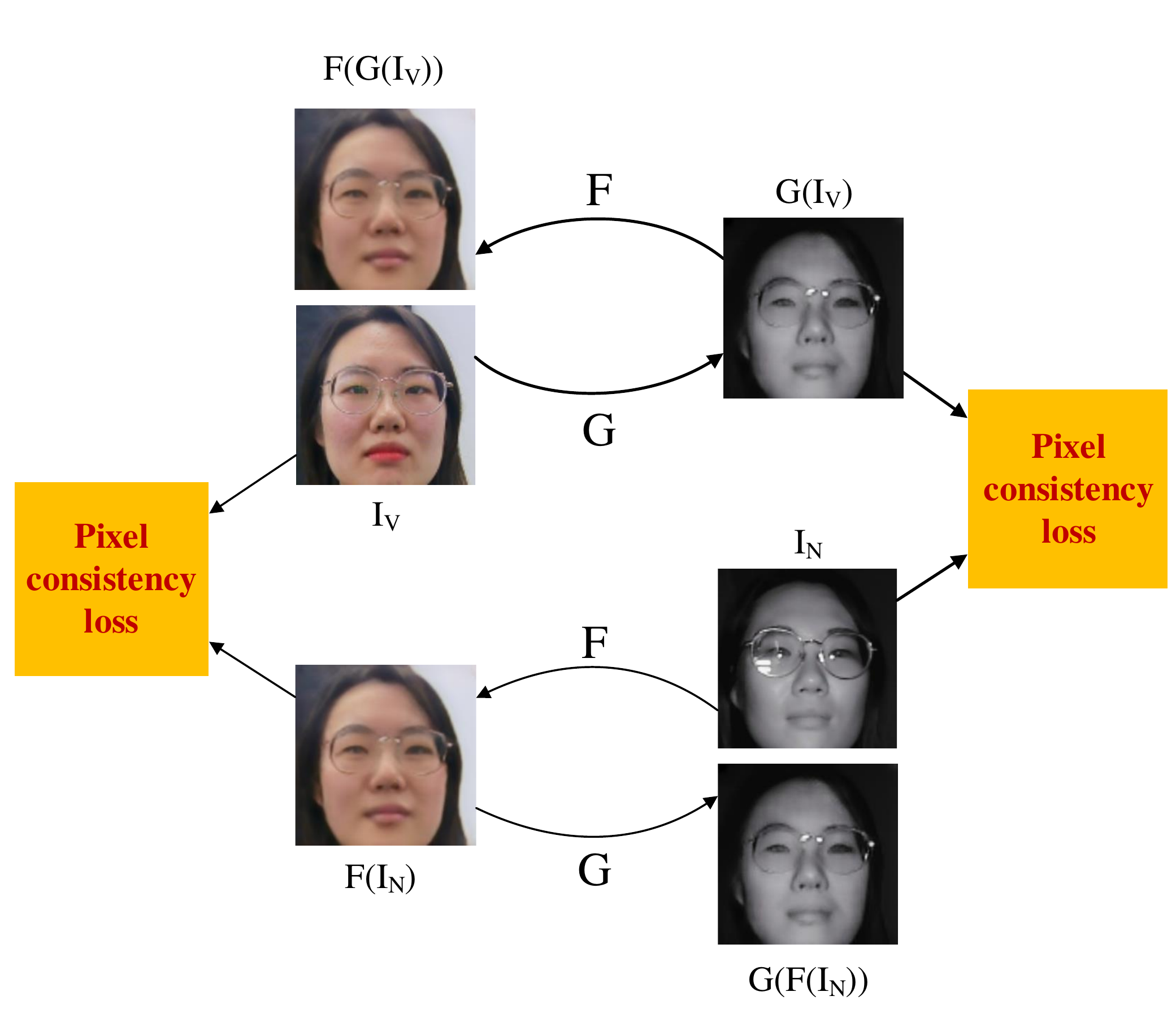}
\end{center}
   \caption{Proposed pixel consistency loss: it enforces the generated fake NIR image $ G(I_{V})$ close to the corresponding real NIR image $ I_{N}$ and vice versa.  }
\label{fig:long}
\label{fig:onecol}
\end{figure}

As shown in Figure 3, we apply a pixel consistency loss to the generated images, which calculate the $L1$ distance between the real images and the generated images from the same domain \cite{ 2017Image-to-Image}. The pixel consistency loss (pc) is proposed to leverage the benefit of our paired VIS-NIR image database. The face images in this database are approximately paired, but not strictly pix-to-pix paired (see Figure 4). Thus we choose the CycleGAN which has more tolerance for the training data (i.e., does not require paired data) as our basic model, and the generated NIR images are restricted by the proposed pc loss, which is formulated as follows:
\begin {equation}
\begin {aligned}
\mathcal{L}_{pc}(G, F) = &\mathbb{E}_{i_{V} \sim  P(i_{V})}\bigg[||G(i_{V})-i_{N}||_1\bigg]  \\
 & + \mathbb{E}_{i_{N} \sim P(i_{N})}\bigg[||F(i_{N})-i_{V}||_1\bigg].
 \end {aligned}
\end {equation}
Therefore, we have the full objective as below:
\begin {equation} 
\begin {aligned}
\mathcal{L}\left(G, F, D_{V}, D_{N}\right) = & \mathcal{L}_{LSGAN}\left(G, D_{N},I_{V}, I_{N}\right) \\
&+\mathcal{L}_{LSGAN}\left(F, D_{V}, I_{N}, I_{V}\right) \\
&+\lambda \mathcal{L}_{cyc}(G, F) + \gamma \mathcal{L}_{pc}(G, F),
\end {aligned}
 \end {equation}
where the parameters $ \lambda $ and $ \gamma $ control the relative importance of different terms.
  

\begin{figure}[t]
\begin{center}
   \includegraphics[width=0.9\linewidth]{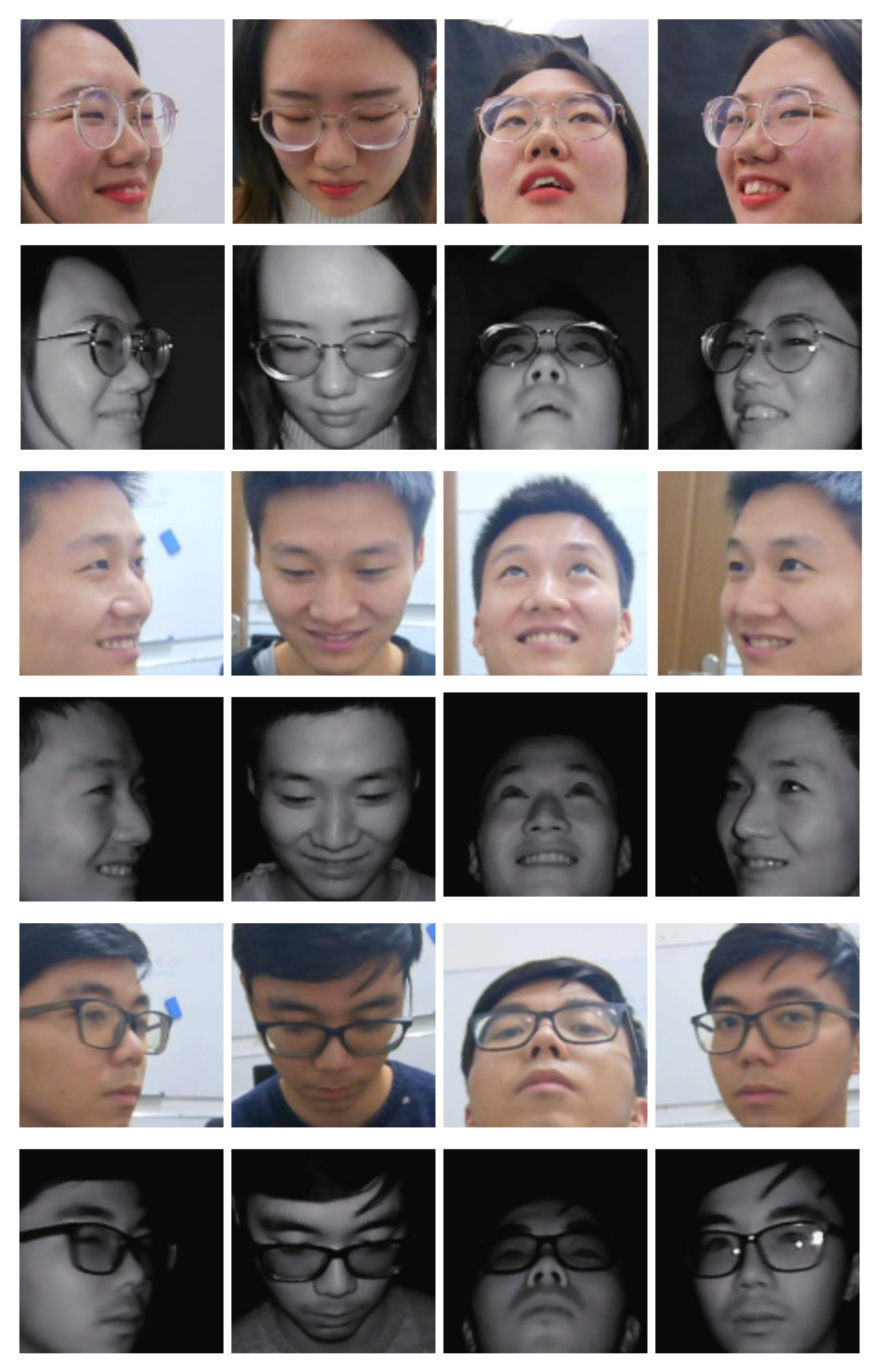}
\end{center}
\caption{Samples from the WHU VIS-NIR paired face database (From left to right: left-rotation, tilt-down, tilt-up,   right-rotation. Every two rows belong to the same subject: VIS image and the corresponding NIR image below).}

\label{fig:long}
\label{fig:onecol}
\end{figure} 


\section{Datasets and Protocol}
In this section, two databases used for performance evaluation are respectively introduced. Firstly, we collect a paired database called the WHU VIS-NIR paired face database.   The videos and images in this database are captured synchronously by a binocular camera (camera model: QR-USB0230X2) in normal indoor illumination. The binocular camera  includes an NIR camera and a VIS camera which capture the same facial expression as shown in Figure 4. This database consists of 80 subjects. Each subject has 2 videos and 80 VIS-NIR paired images, i.e., 80 pictures from VIS domain and 80 pictures from NIR domain. The whole dataset is made up of 12720 images and 160 videos. The images in WHU VIS-NIR database vary in facial rotation and face expressions: neutral-frontal, tilt-up, tilt-down, left-rotation, right-rotation, blank and smile. According to the proposed protocol in \cite{2017TNNLSCross},  we try to select as various poses as possible. Finally, we pick out 20 VIS-NIR pairs, i.e., 20 VIS images and 20 NIR images, for each subject as our training and testing data. Therefore there are totally 3200 images, where 2800 images (1400 VIS images and 1400 NIR images) of 70 subjects make up the training set, and the remaining 400 images (200 VIS and 200 NIR images) of 10 subjects belong to the testing set.  True accepted rates when false accepted rates equal to 1\% and 0.1\% respectively (TAR@FAR=1\%, TAR@FAR=0.1\%), as well as the Rank-1 identification rate are reported as evaluation criteria. 

  
Secondly, the well-known Oulu-CASIA NIR-VIS face expression database \cite{2011PR_oulu-dataset} is also considered for performance evaluation. This database includes six expressions of 80 subjects. Each expression is simultaneously captured by a NIR camera and a VIS camera in three illuminations. Following the method in \cite{2018AAAiADHFR}, we randomly select eight face images in each expression, and only use the normal indoor illumination images. Hence, there are 48 VIS and 48 NIR images from each subject. According to the protocol in \cite{2017TNNLSCross}, the training set and testing set  consist of 20 subjects respectively. So there are totally 960 VIS images and 960 NIR probe images in testing phase. Similarly, TAR@FAR=1\%, TAR@FAR=0.1\%, and Rank-1 identification rate are reported in the experiment section.

\section{Experiments}
\subsection{Experiment Settings} 

We choose the MobileFaceNet \cite{Sheng2018MobileFaceNets} as the feature extraction network, which is  pretrained on MS-Celeb-1M \cite{Guo2016ECCvMSCeleb1M} using Softmax loss and fine-tuned using Arcface loss \cite{Deng2018ArcFace}. The main building blocks of MobileFaceNet consist of the residual bottlenecks proposed in MobileNetV2 \cite{Sheng2018MobileFaceNets}.  All the parameter configurations are the same as those in \cite{Sheng2018MobileFaceNets}. Both of the two generators in CycleGAN use the pretrained MobileFaceNet as the encoder module. The output feature dimension of the encoder is 128, which is delivered to the translate module as an input. The transfer network uses 6 residual blocks and the decoder keeps the same configuration as the original CycleGAN. The proposed FFE-CycleGAN is trained on the WHU VIS-NIR paired face database. We also fine-tune our model for several epochs by using training set from the Oulu-CASIA NIR-VIS face expression database as introduced in Section IV. To prepare face image samples, we use the MTCNN network in \cite{Zhang2016Joint} to detect facial landmarks, based on which face images are aligned and cropped to the size of $ 256 \times 256$. The hyper-parameters $ \lambda $ and $ \gamma $ in Equ. (4) are set to 1 and 10 during training phase.

  
\begin{figure}[t]
\begin{center}
   \includegraphics[width=0.9\linewidth]{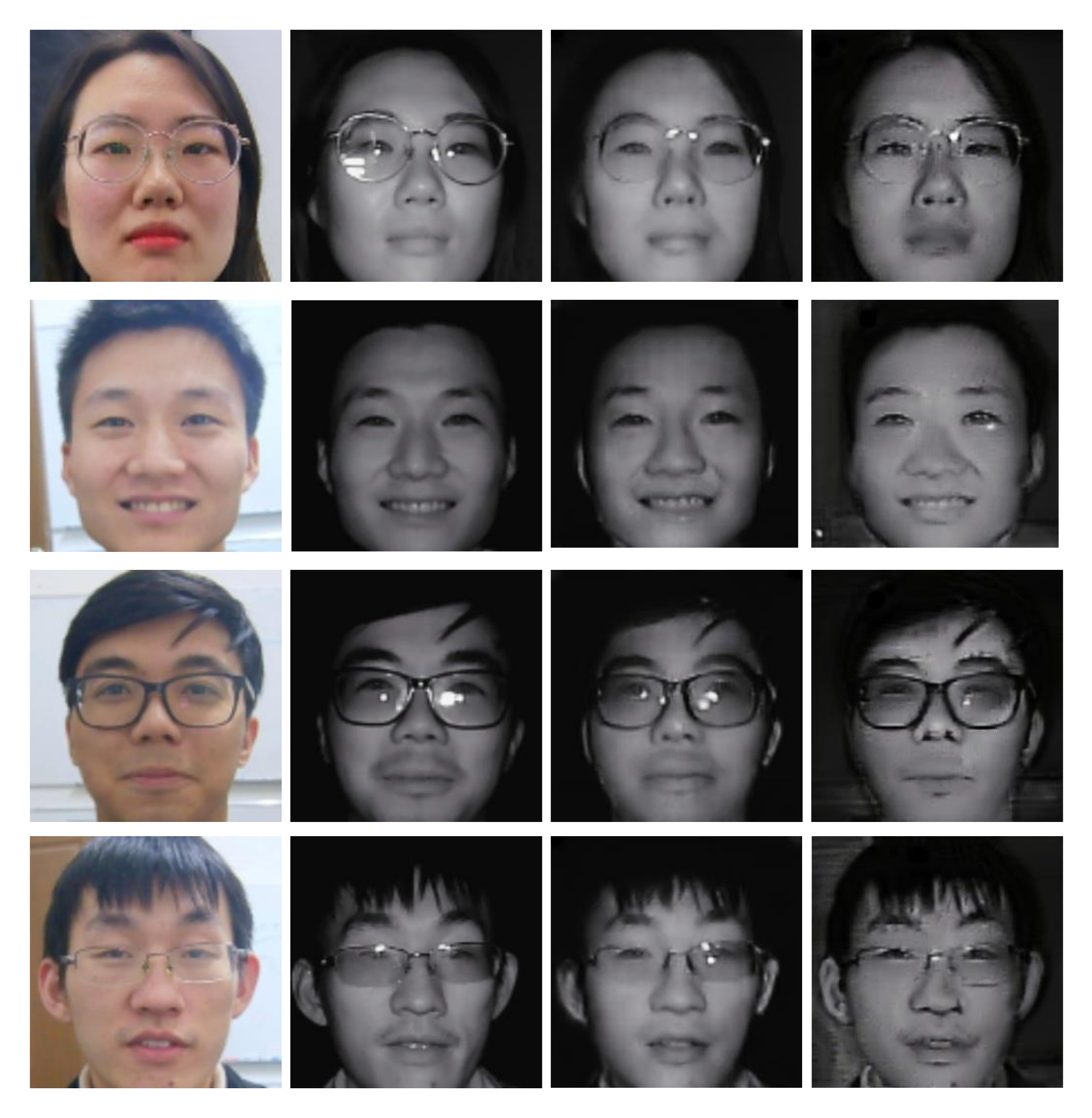}
\end{center}
   \caption{Translated images on the WHU VIS-NIR paired 
face database (From left to right: input VIS images, true NIR images from the same subject, generated fake NIR images by FFE-CycleGAN, and face images by basic CycleGAN model).}
\label{fig:long}
\label{fig:onecol}
\end{figure} 

\begin{table}[t]
\begin{center}
\begin{tabular}{l|c|c|c}
\hline
Method & Rank-1 & TAR@FAR=1\% & TAR@FAR=0.1\% \\   
\hline\hline

Basic CycleGAN & 96.9 & 72.6 & 59.1 \\
Basic + $L_{pc}$  & 97.7 & 73.9 & 59.8 \\
FFE + $L_{pc}$ & \textbf{99.3} & \textbf{76.2} & \textbf{64.0} \\
\hline
\end{tabular}
\end{center}
\caption{Results on WHU VIS-NIR paired face database.}
\end{table}

\subsection{Results and Analysis}


Several translated face images by the proposed FFE-CycleGAN on the WHU VIS-NIR paired face database are shown in Figure 5, where the results of the basic CycleGAN are also presented for comparison. It is seen that the fake NIR images generated by the CycleGAN cannot well learn NIR style feature. CycleGAN seems to simply convert the spectral texture to a VIS image, rather than to consider the variation of face feature, i.e., the shadow of nose and the dark facial outline. Furthermore, the NIR images generated by basic CycleGAN are more artificial, lacking the ray uniformity. The cause of this phenomenon lies in the lack of the proposed constraint $L_{pc}$ for the generated images. More importantly, FFE-CycleGAN  transfers the face feature directly, not the whole picture's feature. Thus, the generated fake NIR images from FFE-CycleGAN look more realistic than images from basic CycleGAN in Figure 5. In contrast, the results of the proposed FFE-CycleGAN in Figure 5 demonstrate that not only the variation of spectral texture can be captured, but also the change from facial features.

Transferring accuracy on the WHU VIS-NIR paired face database is exhibited in Table I in terms of Rank-1, TAR@FAR=1\%, and TAR@FAR=0.1\%. We calculate Rank-1 identification rate,  TAR@FAR=1\%, TAR@FAR=0.1\% for a detailed analysis. We test the basic CycleGAN model \cite{Zhu2017Unpaired} for a fair comparison. To individually validate the effectiveness of the proposed pixel-consistency loss, we train the basic CycleGAN model with $L_{pc}$  (Basic + $L_{pc}$) and report its results in Table I. As we can see, the proposed  $L_{pc}$  boosts the performance of the basic CycleGAN model. In addition, the FFE-CycleGAN  combining with loss $ L_{pc}$ has a significant contribution to improve NIR face recognition accuracy. 

\begin{figure}[t]
\begin{center}
   \includegraphics[width=0.85\linewidth]{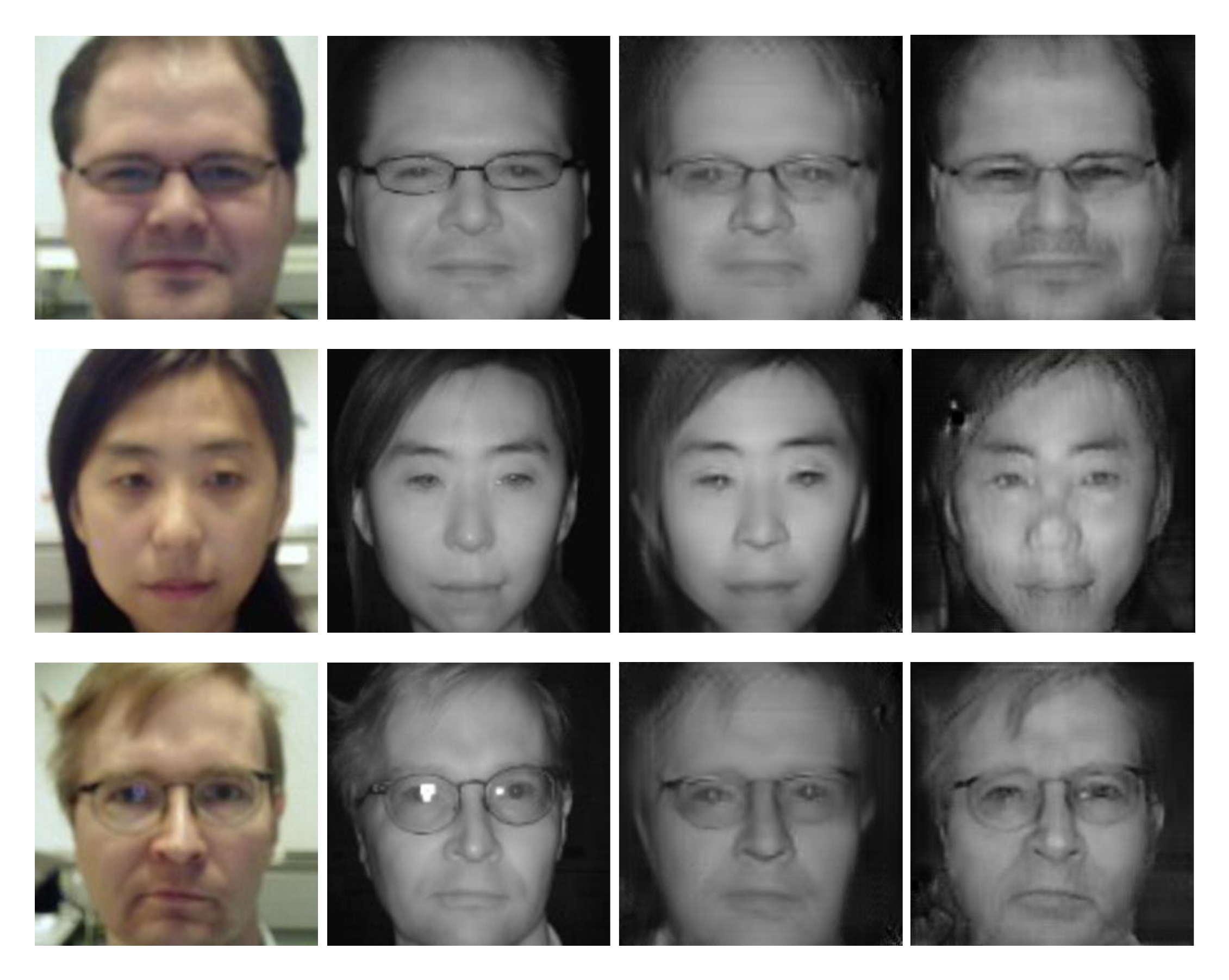}
\end{center}
   \caption{Translated images on Oulu-CASIA NIR-VIS face expression database (From left to right: input VIS images, true NIR images from the same subject, generated fake NIR images by FFE-CycleGAN, and basic CycleGAN).}
\label{fig:long}
\label{fig:onecol}
\end{figure} 

Some of the translated images on Oulu-CASIA NIR-VIS face expression database are shown in Figure 6, as we can see the generated fake NIR images well keep
the facial texture, e.g., the periocular texture. Compared to the images transformed by the basic CycleGAN, the generated images by FFE-CycleGAN look more real and more smooth. The learned common feature shared by VIS and NIR images is the key factor when we do face matching with calculating the verification accuracy and Rank-1 score. Meanwhile, the feature belonging to NIR images is also learned the same as results in Figure 5, like the shadows around nose and mouth. Compared to results converted by the CycleGAN, our fake NIR images look more stereoscopic and well recover the facial texture. Because the FFE-CycleGAN will not only learn the NIR spectrum, but also the face changes, e.g., the dark and blurry outlines caused by the low illumination of NIR, or specific noise distributions of NIR imaging modality. 

We also compare the proposed method with some NIR-VIS face recognition methods, including the mapping learning (MPL) \cite{2009cvprLearningMappings}, kernel couple spectral regression (KCSR) \cite{2009CVPR_Coupled}, kernel prototype similarity (KPS) \cite{2013TPAMI_HFR}, regularized discriminative spectral regression (LDSR/KDSR) \cite{2013TIP_KDSR}, hierarchical hyperlingual-words local binary pattern (LBP) \cite{2009LBP}, $H^2$($LBP_3$){\cite{2017TNNLSCross}, invariant deep representation (IDR) \cite{he2017CVPRlearning}, and adversarial discriminative feature learning (ADFL) \cite{2018AAAiADHFR}. 
Quantitative experiment results on the Oulu-CASIA NIR-VIS facial expression database are exhibited in Table II in terms of Rank-1, TAR@FAR=1\%, and TAR@FAR=0.1\%. 
The experiment results for the above comparing methods are the same as in \cite{2017TNNLSCross}. 
As seen from Table II, in contrast to the basic CycleGAN model, the proposed FFE-CycleGAN model obtains performance gain  at about 3.6\% in Rank-1 accuracy and 2.4\% in TAR@FAR=1\%. In particular, the FFE-CycleGAN model obtains significant performance gain at about 12.8\% in TAR@FAR=0.1\%.  It is worth mentioning that when the $L_{pc}$ is applied, a higher performance gain is achieved. It demonstrates that the $L_{pc}$ can boost the matching accuracy of VIS-NIR images. Besides, the FFE-CycleGAN slightly improves the Rank-1 score and TAR@FAR=0.1\% for about 1.0\% and 0.6\% compared to the ADFL \cite{2018AAAiADHFR} proposed in 2018  which trains on a bigger dataset. This phenomenon demonstrates that the FFE module plays an important role in translating tasks  as analysed in Section III. However, the images generated by our method lose some details during image-to-image transfer, thus it does not further improve the verification rate at high FAR.   As a whole, our proposed network is effective for training on a small number of pairwise face images. And the proposed FFE-CycleGAN performs stably during training, and can translate VIS face images into fake NIR images whose distributions are close to those of true NIR images. 

  
\begin{table}[t]
\begin{center}
\begin{tabular}{l|c|c|c}
\hline
Method & Rank-1 & TAR@FAR=1\% & TAR@FAR=0.1\% \\   
\hline\hline
MPL3(2009) & 48.9 & 41.9 & 11.4\\
KCSR(2009) & 66.0 & 49.7 & 26.1\\
KPS(2013) & 62.2 & 48.3 & 22.2\\
KDSR(2013) & 66.9 & 56.1 & 31.9 \\ 
$H^2$($LBP_3$)(2017) & 70.8 & 62.0 & 33.6\\
IDR(2017) & 94.3 & 73.4 & 46.2\\
ADFL(2018) & 95.5 & \textbf{83.0} & 60.7\\
\hline
Basic CycleGAN & 92.9  & 76.8 & 48.5\\
Basic + $L_{pc}$  & 95.2 & 78.4 & 54.8\\
FFE + $L_{pc}$ & \textbf{96.5} & 79.2 & \textbf{61.3} \\
\hline
\end{tabular}
\end{center}
\caption{Results on Oulu-CASIA NIR-VIS database.}
\end{table}

\section{Conclusion}
This paper proposes a new VIS-NIR translation method by embedding a facial feature extraction network into the original CycleGAN, which ensures effective feature extraction and keeps style translation consistence. The proposed FFE-CycleGAN conducts two strategies through i) combining the MobileFaceNet model with CycleGAN model; and ii) adding a new pixel consistency loss. The experiments validate the correctness and effectiveness of our proposed method for translating a VIS face image into a NIR, which has similar distribution with that of the true NIR image. Consequently, we can match a VIS face image to a NIR image belonging to the same subject by generating a fake NIR image first, and then matching the fake NIR image to the NIR image, which can solve many practical problems under poor lighting conditions. Although the MobileFaceNet is used for extracting face feature, it is feasible to use other facial feature models mentioned in section III. And if there is any bigger paired dataset, welcome to further boost the performance of our network. In future, we will focus on how to further improve translated image qualities, such as higher fidelity and more accurate invariant deep representation between two modalities.

%

\section*{Acknowledgment}
We thank Dr. Guoying Zhao for offering the Oulu-CASIA NIR-VIS face expression database \cite {2011PR_oulu-dataset}. It greatly helps us to further train and test the performance of the proposed method. Besides, we give our great appreciation for those who help us to collect datasets, select pictures and finally build the WHU VIS-NIR paired face database. Meanwhile, we would like to thank the people in Figure 5 and 6 for their generous support. This work was supported in part by National Natural Science Foundation of China under Grant 61802284 and in part by Hubei Provincial Natural Science Foundation of China under grant 2018CFB225.

\ifCLASSOPTIONcaptionsoff
  \newpage
\fi

%
%

\bibliographystyle{IEEEbib}
\bibliography{egbib}

\begin{thebibliography}{10}

\bibitem{Schroff2015FaceNet}
F.~{Schroff}, D.~{Kalenichenko}, and J.~{Philbin},
\newblock ``{FaceNet}: A unified embedding for face recognition and
  clustering,''
\newblock in {\em 2015 IEEE Conference on Computer Vision and Pattern
  Recognition (CVPR)}, June 2015, pp. 815--823.

\bibitem{Sheng2018MobileFaceNets}
S.~Chen, Y.~Liu, X.~Gao, and Z.~Han,
\newblock ``{MobileFaceNets}: Efficient {CNNs} for accurate real-time face
  verification on mobile devices,''
\newblock {\em CoRR}, vol. abs/1804.07573, 2018.

\bibitem{ouyang2016survey}
S.~Ouyang, T.~Hospedales, Y.~Song, X.~Li, C.~Loy, and X.~Wang,
\newblock ``A survey on heterogeneous face recognition: Sketch, infra-red, {3D}
  and low-resolution,''
\newblock {\em Image and Vision Computing}, vol. 56, pp. 28--48, 2016.

\bibitem{he2017CVPRlearning}
R.~He, X.~Wu, Z.~Sun, and T.~Tan,
\newblock ``Learning invariant deep representation for {NIR-VIS} face
  recognition.,''
\newblock in {\em AAAI Conference on Artificial Intelligence}, 2017, vol.~4,
  p.~7.

\bibitem{gao2017semi}
Y.~Gao, J.~Ma, and A.~L. Yuille,
\newblock ``Semi-supervised sparse representation based classification for face
  recognition with insufficient labeled samples,''
\newblock {\em IEEE Transactions on Image Processing}, vol. 26, no. 5, pp.
  2545--2560, 2017.

\bibitem{Zhu2017Matching}
J.~Zhu, W.~Zheng, J.~Lai, and S.~Li,
\newblock ``Matching {NIR} face to {VIS} face using transduction,''
\newblock {\em IEEE Transactions on Information Forensics \& Security}, vol. 9,
  no. 3, pp. 501--514, 2017.

\bibitem{2018AAAiADHFR}
L.~Song, M.~Zhang, X.~Wu, and R.~He,
\newblock ``Adversarial discriminative heterogeneous face recognition,''
\newblock in {\em AAAI Conference on Artificial Intelligence}, 2018.

\bibitem{Zhu2017Unpaired}
J.~{Zhu}, T.~{Park}, P.~{Isola}, and A.~A. {Efros},
\newblock ``Unpaired image-to-image translation using cycle-consistent
  adversarial networks,''
\newblock in {\em 2017 IEEE International Conference on Computer Vision
  (ICCV)}, Oct 2017, pp. 2242--2251.

\bibitem{2011PR_oulu-dataset}
G.~{Zhao}, X.~Huang, M.~{Taini}, S.~Z. {Li}, and M.~{Pietik\" ainen},
\newblock ``Facial expression recognition from near-infrared videos,''
\newblock {\em Image and Vision Computing}, vol. 29, pp. 607--619, Aug 2011.

\bibitem{2009cvprLearningMappings}
J.~Chen, D.~{Yi}, J.~Yang, G.~Zhao, S.~Z. {Li}, and M.~{Pietik\" ainen},
\newblock ``Learning mappings for face synthesis from near infrared to visual
  light images,''
\newblock in {\em 2009 IEEE Conference on Computer Vision and Pattern
  Recognition}, June 2009, pp. 156--163.

\bibitem{2009CVPR_Coupled}
Z.~Lei and S.~Z. {Li},
\newblock ``Coupled spectral regression for matching heterogeneous faces,''
\newblock in {\em 2009 IEEE Conference on Computer Vision and Pattern
  Recognition}, June 2009, pp. 1123--1128.

\bibitem{2013TPAMI_HFR}
B.~F. {Klare} and A.~K. {Jain},
\newblock ``Heterogeneous face recognition using kernel prototype
  similarities,''
\newblock {\em IEEE Transactions on Pattern Analysis and Machine Intelligence},
  vol. 35, no. 6, pp. 1410--1422, June 2013.

\bibitem{2013TIP_KDSR}
X.~{Huang}, Z.~{Lei}, M.~{Fan}, X.~{Wang}, and S.~Z. {Li},
\newblock ``Regularized discriminative spectral regression method for
  heterogeneous face matching,''
\newblock {\em IEEE Transactions on Image Processing}, vol. 22, no. 1, pp.
  353--362, Jan 2013.

\bibitem{2017TNNLSCross}
M.~{Shao} and Y.~{Fu},
\newblock ``Cross-modality feature learning through generic hierarchical
  hyperlingual-words,''
\newblock {\em IEEE Transactions on Neural Networks and Learning Systems}, vol.
  28, no. 2, pp. 451--463, 02 2017.

\bibitem{2017CVPRNotAfraidofDark}
J.~Lezama, Q.~Qiu, and G.~Sapiro,
\newblock ``Not afraid of the dark: {NIR-VIS} face recognition via
  cross-spectral hallucination and low-rank embedding,''
\newblock in {\em 2017 IEEE Conference on Computer Vision and Pattern
  Recognition (CVPR)}. IEEE, 2017, pp. 6807--6816.

\bibitem{2015CVPR_NIR-VIS}
F.~Juefei-Xu, D.~K. Pal, and M.~Savvides,
\newblock ``{NIR-VIS} heterogeneous face recognition via cross-spectral joint
  dictionary learning and reconstruction,''
\newblock in {\em 2015 IEEE Conference on Computer Vision and Pattern
  Recognition Workshops (CVPRW)}, June 2015, pp. 141--150.

\bibitem{Wang2009Analysis-by-Synthesis}
R.~Wang, J.~Yang, D.~Yi, and S.~Z. Li,
\newblock ``An analysis-by-synthesis method for heterogeneous face
  biometrics,''
\newblock in {\em Advances in Biometrics}, Massimo Tistarelli and Mark~S.
  Nixon, Eds., Berlin, Heidelberg, 2009, pp. 319--326, Springer Berlin
  Heidelberg.

\bibitem{2016ICB_Transferring}
X.~Liu, L.~Song, X.~{Wu}, and T.~{Tan},
\newblock ``Transferring deep representation for {NIR-VIS} heterogeneous face
  recognition,''
\newblock in {\em 2016 International Conference on Biometrics (ICB)}, 06 2016,
  pp. 1--8.

\bibitem{2016ECCV_HFRwithCNN}
S.~Saxena and J.~Verbeek,
\newblock ``Heterogeneous face recognition with {CNNs},''
\newblock in {\em Computer Vision -- ECCV 2016 Workshops}, Gang Hua and
  Herv{\'e} J{\'e}gou, Eds., Cham, 2016, pp. 483--491, Springer International
  Publishing.

\bibitem{2016cvpr_Seeing}
C.~{Reale}, N.~M. {Nasrabadi}, H.~{Kwon}, and R.~{Chellappa},
\newblock ``Seeing the forest from the trees: A holistic approach to
  near-infrared heterogeneous face recognition,''
\newblock in {\em 2016 IEEE Conference on Computer Vision and Pattern
  Recognition Workshops (CVPRW)}, June 2016, pp. 320--328.

\bibitem{2017arxiv_Coupled}
X.~Wu, L.~Song, R.~He, and T.~Tan,
\newblock ``Coupled deep learning for heterogeneous face recognition,''
\newblock {\em CoRR}, vol. abs/1704.02450, 2017.

\bibitem{Li2016ACM_Mutual}
Z.~Li, D.~Gong, Q.~Li, D.~Tao, and X.~Li,
\newblock ``Mutual component analysis for heterogeneous face recognition,''
\newblock {\em ACM Trans. Intell. Syst. Technol.}, vol. 7, no. 3, pp.
  28:1--28:23, Mar. 2016.

\bibitem{Goodfellow2014NIPSGAN}
I.~Goodfellow, J.~Pouget-Abadie, M.~Mirza, X.~Bing, D.~Warde-Farley, S.~Ozair,
  A.~Courville, and Y.~Bengio,
\newblock ``Generative adversarial nets,''
\newblock in {\em International Conference on Neural Information Processing
  Systems}, 2014.

\bibitem{2017Image-to-Image}
P.~Isola, J.~Zhu, T.~Zhou, and A.~A. Efros,
\newblock ``Image-to-image translation with conditional adversarial networks,''
\newblock in {\em 2017 IEEE Conference on Computer Vision and Pattern
  Recognition (CVPR)}, July 2017, pp. 5967--5976.

\bibitem{2015DeepID3}
Y.~Sun, D.~Liang, X.~Wang, and X.~Tang,
\newblock ``{DeepID3}: Face recognition with very deep neural networks,''
\newblock {\em CoRR}, vol. abs/1502.00873, 2015.

\bibitem{Sandler2018MobileNetV2}
M.~{Sandler}, A.~{Howard}, M.~{Zhu}, A.~{Zhmoginov}, and L.~{Chen},
\newblock ``{MobileNetV2}: Inverted residuals and linear bottlenecks,''
\newblock in {\em 2018 IEEE/CVF Conference on Computer Vision and Pattern
  Recognition}, June 2018, pp. 4510--4520.

\bibitem{Guo2016ECCvMSCeleb1M}
Y.~Guo, L.~Zhang, Y.~Hu, X.~He, and J.~Gao,
\newblock ``{MS-Celeb-1M}: A dataset and benchmark for large-scale face
  recognition,''
\newblock in {\em Computer Vision -- ECCV 2016}, Bastian Leibe, Jiri Matas,
  Nicu Sebe, and Max Welling, Eds., Cham, 2016, pp. 87--102, Springer
  International Publishing.

\bibitem{mao2016multi}
X.~Mao, Q.~Li, H.~Xie, R.~Y. Lau, and Z.Wang,
\newblock ``Multi-class generative adversarial networks with the {L2} loss
  function,''
\newblock {\em arXiv preprint arXiv:1611.04076}, vol. 5, 2016.

\bibitem{Deng2018ArcFace}
J.~Deng, J.~Guo, N.~Xue, and S.~Zafeiriou,
\newblock ``Arcface: Additive angular margin loss for deep face recognition,''
\newblock {\em arXiv preprint arXiv:1801.07698}, 2018.

\bibitem{Zhang2016Joint}
K.~Zhang, Z.~Zhang, Z.~Li, and Q.~Yu,
\newblock ``Joint face detection and alignment using multitask cascaded
  convolutional networks,''
\newblock {\em IEEE Signal Processing Letters}, vol. 23, no. 10, pp.
  1499--1503, 2016.

\bibitem{2009LBP}
S.~Liao, D.~Yi, Z.~Lei, R.~Qin, and S.~Z. Li,
\newblock ``Heterogeneous face recognition from local structures of normalized
  appearance,''
\newblock in {\em Advances in Biometrics}, Massimo Tistarelli and Mark~S.
  Nixon, Eds., Berlin, Heidelberg, 2009, pp. 209--218, Springer Berlin
  Heidelberg.

\end{thebibliography}

%
%
%

\end{document}